\documentclass[10pt,twocolumn,letterpaper]{article}

\usepackage{cvpr}
\usepackage{times}
\usepackage{epsfig}
\usepackage{graphicx}
\usepackage{amsmath}
\usepackage{amssymb}

\usepackage{booktabs}
\usepackage{multirow}
\usepackage{makecell}
\usepackage{arydshln}

\usepackage[ruled]{algorithm2e}
\usepackage{enumitem}
\usepackage{color,soul}
\usepackage{wrapfig}

\DeclareMathOperator*{\argmin}{arg\,min}

\usepackage[pagebackref=true,breaklinks=true,letterpaper=true,colorlinks,bookmarks=false]{hyperref}

\cvprfinalcopy 

\ifcvprfinal\fi
\begin{document}

\title{DMCL: Distillation Multiple Choice Learning \\
for Multimodal Action Recognition}
    

\author{
Nuno Cruz Garcia$^{1,2}$ \quad\quad\quad Sarah Adel Bargal$^3$ \quad\quad\quad Vitaly Ablavsky$^3$ \\
 Pietro Morerio$^1$ \quad\quad\quad\quad \  Vittorio Murino$^{1,4,5}$ \quad\quad\quad Stan Sclaroff$^3$ \vspace{.5em} \\
$^1$Istituto Italiano di Tecnologia  \quad $^2$Universit\`a di Genova \\
\quad $^3$Boston University \quad $^4$Universit\`a di Verona \quad $^5$Huawei Ireland Research Center \\
{\tt\small \{nuno.garcia,pietro.morerio,vittorio.murino\}@iit.it, \{sbargal,ablavsky,sclaroff\}@bu.edu} \vspace{-.5em}
}

\definecolor{blue(ryb)}{rgb}{0.01, 0.28, 1.0}
\definecolor{ao(english)}{rgb}{0.0, 0.5, 0.0}
\definecolor{burntorange}{rgb}{0.8, 0.33, 0.0}
\newcommand\boldrgb[1]{\textcolor{blue(ryb)}{\textbf{#1}}}
\newcommand\boldflow[1]{\textcolor{burntorange}{\textbf{#1}}}
\newcommand\bolddepth[1]{\textcolor{ao(english)}{\textbf{#1}}}
\newcommand\rgb[1]{\textcolor{blue(ryb)}{#1}}
\newcommand\flow[1]{\textcolor{burntorange}{#1}}
\newcommand\depth[1]{\textcolor{ao(english)}{#1}}
\newcommand\urgb[1]{\setulcolor{blue(ryb)}{\ul{#1}}}
\newcommand\uflow[1]{\setulcolor{burntorange}{\ul{#1}}}
\newcommand\udepth[1]{\setulcolor{ao(english)}{\ul{#1}}}

\let\oldnl\nl
\newcommand{\nonl}{\renewcommand{\nl}{\let\nl\oldnl}}

\setlength{\abovedisplayskip}{3pt}
\setlength{\belowdisplayskip}{3pt}
\setlength{\abovedisplayshortskip}{3pt}
\setlength{\belowdisplayshortskip}{3pt}

\maketitle

\begin{abstract}
In this work, we address the problem of learning an ensemble of specialist networks using multimodal data, while considering the realistic and challenging scenario of possible missing modalities at test time.
Our goal is to leverage the complementary information of multiple modalities to the benefit of the ensemble and each individual network.
We introduce a novel Distillation Multiple Choice Learning framework for multimodal data, where different modality networks learn in a cooperative setting from scratch, strengthening one another. 
The modality networks learned using our method achieve significantly higher accuracy than if trained separately, due to the guidance of other modalities. We evaluate this approach on three video action recognition benchmark datasets. We obtain state-of-the-art results in comparison to other approaches that work with missing modalities at test time.
\end{abstract}

\section{Introduction}
Humans perceive the environment by processing a combination of modalities. Such modalities can include audio, touch and sight, with each modality being distinct from and complementary to the others. 
Deep learning methods may likewise benefit from multimodal data.
In this paper, we explore how to leverage the complementary nature of multimodal data at training time, in order to learn a better classifier that takes as input only RGB data for inference.

One popular way to train multimodal deep learning models 
is to train one network per modality, and mean pool all the network predictions for inference. 
This is a sub-optimal use of multimodal training data, as modalities do not exchange information while training.
For example, considering the task of action recognition, some actions are easier to discriminate using certain modalities over others: 
the action ``open a box" may be confused with ``fold paper" when solely relying on the RGB modality, while it is easily classified using depth data \cite{liu2019ntu}.

\begin{figure}[t]
    \centering
    \includegraphics[width=\linewidth,trim={0.0in 0.0in 0.0in 0.0in},clip]{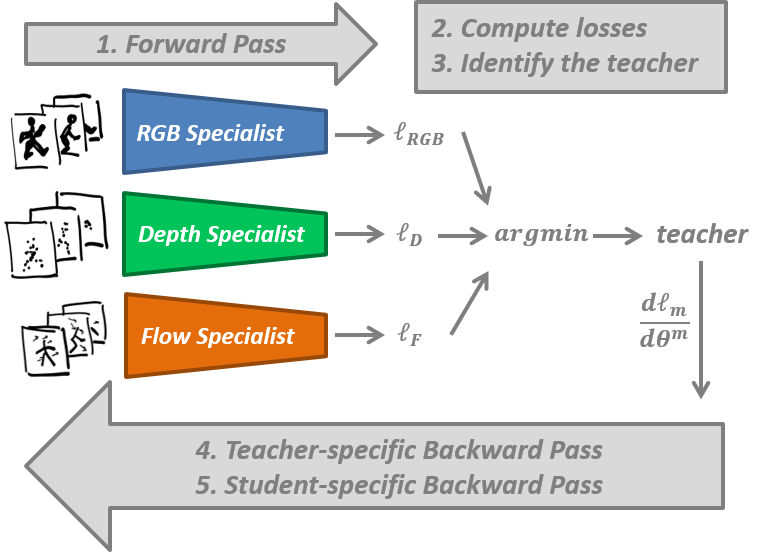}
    \vspace{-4mm}
\caption{\footnotesize \textbf{Distillation Multiple Choice Learning (DMCL)} allows multiple modalities to cooperate and strengthen one another. For each training sample, the modality specialist $m$ that achieves the lowest loss $\ell$ distills knowledge to strengthen other modality specialists.
At test time, any subset of available modalities can be used by DMCL to make predictions.}
\label{fig:specialists}
\end{figure}

This suggests that an ensemble of networks could use multimodal data in a more efficient way, \textit{e.g.} by encouraging the network trained with a given modality to focus on the set of classes or samples that maximizes its discriminative power.
In this case, each network is referred to as a \textit{specialist network}, as it only sees part of the dataset and specializes in that part of the problem. 
Assuming that all modalities are available, the ensemble should be able to fuse the specialists' predictions and produce a single output.

The problem of multimodal fusion becomes more challenging when some modalities are not available at test time.
This is particularly problematic if the training process encourages the specialization of each modality network of the ensemble.
In this case, a missing modality means that the ensemble loses the ability to correctly classify the corresponding part of the task assigned to this specialist.

In this paper, we propose a novel method that is at the intersection of MCL framework and Knowledge Distillation \cite{hinton2015distilling,lopez2015unifying}, called Distillation Multiple Choice Learning (DMCL). DMCL addresses two practical dimensions of multimodal learning: a) leveraging the complementarity of multiple modalities, and b) being robust to missing modalities at test-time. 

We take inspiration from the Multiple Choice Learning (MCL) framework, which is a popular way to train an ensemble of RGB networks 
\cite{lee2016stochastic,lee2017confident,tian2019versatile}. This method chooses the best performing network of the ensemble to backpropagate the task loss.
However, extending it to multiple modalities is not straightforward.
Networks that are trained using different modalities learn at different speeds. 
 Consequently, the network that learns faster in the beginning of the training dominates the traditional MCL algorithm, and is encouraged to remain dominant throughout the training.
We extend MCL to a) address such challenges associated with multimodal data, and b) deal with modalities that may be missing at test time.

The case of a missing modality at test time is related to learning using Privileged Information \cite{vapnik2009new} and Knowledge Distillation \cite{hinton2015distilling}. 
This type of approaches is usually structured as a two-step process: training a teacher network, and then using its knowledge to train a student network. The teacher network has usually a larger capacity, or has access to more data than the student.
For example, consider the problem of learning a model for action recognition using a multimodal dataset composed of RGB, depth, and optical flow videos.
In practice, it is reasonable to assume that only RGB modality is present for test inference: depth sensors are expensive and optical-flow computation incurs runtime cost that may not meet real-time budget.
At the same time, depth and optical flow can provide valuable information on the samples or classes that it perform better, and that could be distilled to the RGB network \cite{simonyan2014two} \cite{Carreira_2017_CVPR}.

We build on these ideas to develop a model that learns from multimodal data, exploiting the strength of each modality in a cooperative setting as the training proceeds. This is summarized in Figure~\ref{fig:specialists}. Furthermore, our proposed model is able to account for one or more missing modalities at test time. The code of our Tensorflow \cite{abadi2016tensorflow} implementation is available at \url{https://github.com/ncgarcia/DMCL}.
Our main contributions are:
\begin{itemize}[noitemsep,topsep=1pt]
    \item We conduct a deep evaluation of the MCL framework in the context of multimodal learning and give insights on how multiple modalities behave in such ensemble learning methods.
    \item We propose DMCL, a MCL framework designed for multimodal data where modalities cooperate to strengthen one another.
    Moreover, DMCL is able to account for missing modalities at test time.
    \item We present competitive to or state-of-art results for multimodal action recognition using privileged information on three video action recognition benchmark datasets.
\end{itemize}

\begin{figure*}[t!]
    \centering
    \includegraphics[width=\textwidth,trim={0.0in 0.0in 0.0in 0.0in},clip]{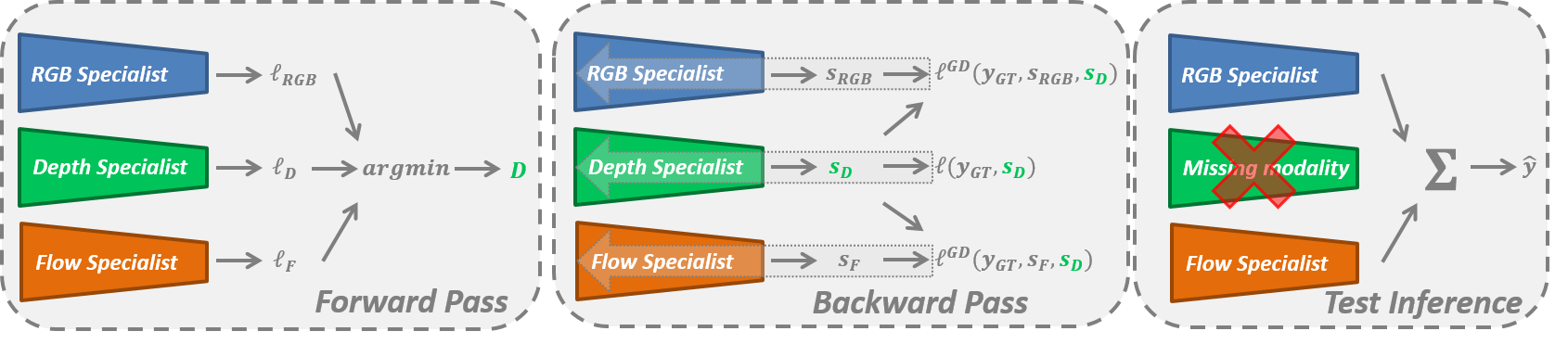}
     \vspace{-5mm}
\caption{\footnotesize \textbf{Distillation Multiple Choice Learning (DMCL)} In the Forward Pass, we calculate the classification cross-entropy losses $\ell$ for each modality and identify the teacher network - in this case, the Depth network. In the Backward Pass, we compute the soft targets of the teacher, $S_D$, and use them as an extra supervision signal for the student networks. The loss for the student networks $\ell^{GD}$ refers to the Generalized Distillation loss, defined on Eq. \ref{eq:loss_GD}. The loss for the teacher network $D$ uses the normal logits, \ie soft targets with temperature $T=1$. At test time, we are able to cope with missing modalities. The final prediction is obtained by averaging the predictions of the available modalities.}
\label{fig:method}
\end{figure*}

\section{Related Work}
\label{RW}
\noindent \textbf{Generalized Distillation.} The Generalized Distillation \cite{lopez2015unifying} framework gives a unifying perspective on Knowledge Distillation (KD) \cite{hinton2015distilling} and Learning Using Privileged Information (LUPI) \cite{vapnik2009new}.
KD was first proposed as a way to transfer knowledge from a large ensemble of networks to a single small capacity network \cite{hinton2015distilling}.
It uses the smoothed ensemble's probability distribution as a soft target to train the lighter network, in addition to the ground truth target.
LUPI refers to the setting where some information available at training time is not be available at test time \cite{vapnik2009new}.
The privileged information can be provided by a "teacher" network, for example, a model previously trained on another dataset or modalities.
The "student" network leverages the additional information to learn a better model to be used at test time.

These ideas have been applied in many creative ways to a variety of domains such as network compression \cite{bucilua2006model}, language tasks \cite{furlanello2018born}, defending from adversarial attacks \cite{papernot2016distillation}, transfer labels across domains \cite{gupta2016cross}, unifying classifiers using unlabeled data \cite{vongkulbhisal2019unifying}, or using distillation without a pre-trained teacher \cite{zhang2018deep} \cite{yang2019snapshot}.
Inspired by these ideas, we extend the MCL algorithm for multimodal tasks, allowing knowledge transfer between modalities in a cooperative learning setting via KD. 

\smallskip

\noindent \textbf{Video Action Recognition.} 
Video action recognition has a vast body of literature.
We focus on multimodal deep learning methods in a privileged information setting, \ie using fewer modalities at test time.
A more comprehensive review is presented in \cite{wang2018rgb} \cite{herath2017going} \cite{kong2018human}.
The combination of RGB and Optical Flow is one of the most popular ways to capture appearance and temporal information for video tasks \cite{simonyan2014two}.
Some interesting works use modules specifically developed to learn motion features, which are then incorporated in models that use RGB only \cite{sun2018optical} \cite{lee2018motion} \cite{piergiovanni2019representation} \cite{zhao2019dance}. Due to the specificity of these modules, these architectures can be difficult to adapt to incorporate other kind of features or modalities, such as depth.
Other works used distillation to transfer knowledge across modalities. 
For example Luo \etal \cite{Luo_2018_ECCV} proposes a graph mechanism to mediate the strength of the imitation loss between modalities. Other methods learn an additional hallucination network to mimic the features of a missing modality \cite{Garcia_2018_ECCV} \cite{garcia2018learning} \cite{crasto2019mars}.
These works use all data of all modalities indiscriminately, and learning the additional hallucination network requires a pre-trained network. Our method learns by exploring the multimodal data asymmetrically via the MCL algorithm, which leverages the strengths of each modality, without the need of a pre-training step or an additional network at test time.

\smallskip

\noindent \textbf{Ensemble Methods.} 
A comprehensive review about ensemble methods is well presented in \cite{sagi2018ensemble}. The most relevant method to ours is the Multiple Choice Learning (MCL) framework. Guzman-Rivera \etal \cite{guzman2012multiple} proposed MCL to optimize the oracle accuracy of an ensemble of models. Lee \etal \cite{lee2016stochastic} proposed Stochastic MCL, an adaptation of MCL to an ensemble of neural networks that learn via stochastic gradient descent. 
Each network of the ensemble trained via Stochastic MCL produces a set of diverse outputs. The inability to output a single prediction compromises its use in real applications. Lee \etal \cite{lee2017confident} addressed this issue with Confident MCL. The main idea is to avoid confident predictions for the classes not assigned to a given specialist. This allows for the sum of all ensemble's networks outputs to get a single prediction. Tian \etal \cite{tian2019versatile} also addressed this issue by training an additional network to estimate the weight of the outputs of each specialist.
While \cite{lee2017confident} and \cite{tian2019versatile} propose ways to get a single prediction out of the ensemble, they do not address how such methods can be used with multimodal data. 
We draw inspiration on these works to address this issue within the MCL framework.

\section{Method: Training Multimodal Specialists}
\label{sec:method}
Our goal is to learn an ensemble of multimodal specialists that leverages the specific strengths of each modality to the benefit of the ensemble. This is accomplished by setting a cooperative learning strategy where stronger networks teach weaker networks through knowledge distillation. For a given data point at training time, we identify the best-performing network as a teacher for the remaining networks in the ensemble.

\smallskip

\subsection{Distillation Multiple Choice Learning} 
Algorithm \ref{alg:gmcl} describes our method DMCL. 
Let ${\mathbb{D}=\{(x_i, y_i)\}^N}$ be a multimodal dataset having $N$ training samples. Each sample $x_i$ represents the data for the $M$ modalities available, ${x_i=\{x_i^1,\dots,x_i^M\}}$, and $y_i$ represents its label.

Our ensemble is composed of a set of $M$ networks $f$, each using as input a different modality $f^1(x_i^1),\dots,f^M(x_i^M))$.
The MCL algorithm maximizes the ensemble accuracy, often referred to as oracle accuracy. The oracle accuracy assumes that we can choose the correct prediction out of the set of outputs produced by each network.
This translates to the minimization of the ensemble loss $L$, which is defined as the lowest of the individual networks' loss values, calculated for a given data point. 

Formally, MCL minimizes the ensemble loss $L$ with respect to a specific task loss $\ell(y_i,\hat{y_i})$ for each network prediction $\hat{y_i}=f^m(x_i^m)$ for a specific modality $m$:
\begin{equation}
L(\mathbb{D}) = \sum \limits_{i=1}^N  \min \limits_{m\in \{1, \dots, M\}} \ell (y_i, f^m(x_i^m)).
\label{eq:oracleloss}
\end{equation}

In practice, we get all the networks' predictions for each sample of the batch. We calculate the loss $\ell_{criterion}$ for each network and sample (line 5, Algorithm~\ref{alg:gmcl}). In this case, $\ell_{criterion}$ corresponds to the standard cross-entropy loss.
The network with the lowest loss value is designated as the winner network, and the others are set to be loser networks. The loss and gradient updates for a network depend on whether it is a winner or loser network (lines 10-14, Algorithm~\ref{alg:gmcl}). In our proposed privileged-information formulation, we view the winner network as a teacher, and the loser networks as students. 

DMCL function of \FuncSty{update\_winner} and \FuncSty{update\_losers} of Algorithm~\ref{alg:gmcl} define how the teacher network distills information to the student networks, strengthening them. DMCL updates teachers with respect to the cross-entropy training loss computed using the ground-truth label. The loser networks are updated using a distillation loss, which aims to transfer knowledge from the winner network. 

\smallskip

\noindent \textbf{Knowledge Distillation.} Matching the students' with the teachers' soft targets is one way to transfer knowledge from one model to another. 
Soft targets are a smoothed probability distribution than the originally produced by the modality network $f^m$:
\begin{equation}
\label{eq:soft_targ}
s_i^m = \sigma(f^m_i(x_i^m) / T),
\end{equation}
where $\sigma$ is the softmax function, $f^m_i$ are the logits, and $T$ is a scalar value. The default temperature $T$ value is set to 1 for models that do not incorporate distillation. Setting $T$ to a higher value produces a smoother probability distribution that reveals valuable information about the relative probabilities between classes, which has shown to improve knowledge transfer and generalization of the new model. 
In practice, very small probability values become more evident with higher temperatures.

The Generalized Distillation (GD) \cite{lopez2015unifying} method consists of three sequential steps: (1) learn the teacher network; (2) fix the teacher and compute the soft target for all samples; (3) use the teacher's soft targets as additional targets to the ground truth to learn student networks. The Generalized Distillation loss is defined as:
\begin{equation}
\label{eq:loss_GD}
\begin{split}
    \ell^{GD}(i) = (1-\lambda) & \ell (y_i,\sigma(f(x_i))) \\
    + \lambda & \ell (s_i, \sigma (f(x_i))), \lambda \in [0,1]
\end{split}
\end{equation}
In contrast, we use distillation in an online fashion in the context of the MCL framework. The role of teacher / student network is assigned to the winner / loser network respectively, for each sample of the batch. The soft targets are computed using the winner network output, which is used to compute the loss and update the loser networks. We do not pretrain teachers as per conventional distillation, \ie all networks are randomly initialized. In DMCL, teachers and students learn together in a cooperative setting.

This cooperative setting is beneficial in two ways: It gives loser networks the opportunity to build good representations even if they are not the \textit{argmin} chosen network ; 
It still enables networks to specialize in parts of the problem.

\smallskip

\noindent \textbf{Missing modalities.} Our training method encourages each network to learn using ground truth labels for its specialty samples (those obtaining lowest loss), and from the other specialist networks for samples otherwise. By doing so, each specialist incorporates knowledge related to all samples/classes of the task. This enables each network to classify any sample at test time, therefore rendering the ensemble able to account for missing modalities.

\begin{algorithm}[t]
\LinesNumbered
\KwIn{Dataset ${\mathbb{D}=\{(x_i, y_i)\}_i^N}$, and randomly initialized networks $f^1, \dots, f^M$ parameterized by $\theta^1,\dots,\theta^M$}
\KwOut{$M$ trained networks $f^1, \dots, f^M$}
\For{step $\gets 1$ to convergence} {
  Sample batch $\mathbb{B} \subset \mathbb{D}$ \\ 
  \For{$m \gets 1$ to $M$}{
Forward Pass: \\
$\ell_{criterion}^m =$ cross\_entropy ($y_i, \hat{y}^m)$\\
  }
\For{$i \gets 1$ to $|\mathbb{B}|$} {
  \textit{// Backward Pass:} \\
  \textit{// Update winner network $m^*$ } \\
  $m^* \gets \argmin\limits_{m \in \{1,\dots,M\}} \{\ell_{criterion}^m\}$ \\ 
  $\theta^{m^*} = $\FuncSty{update\_winner}~($\theta^{m^*}$, $x_i^{m^*}$, $y_i$, $f$) \\ 
  \textit{// Update loser networks $m^c$ } \\
  $m^c \gets \{1,...,M\} \setminus \{m^*\}$ \\
  $\theta^{m^c} =  \FuncSty{update\_losers} $ ~($\theta^{m^c}$, $x_i^{m^c}$, $y_i$, $f$)
}
}
\Return{$f^1 ,$  \dots $,f^M$} \\
\vspace{0.5em}
// \textbf{\textit{Function Definitions}} \\  
\vspace{0.5em}
\SetKwFunction{FWinner}{update\_winner}
 \SetKwFunction{FLoser}{update\_losers}
\SetKwProg{Fn}{Function}{:}{}
 \Fn{\FWinner{$\theta^{m^*}$, $x_i^{m^*}$, $y_i$, $f$}}{
        \textit{// Compute the gradient w.r.t. cross-entropy loss}\;
        $\nabla_{\theta^{m^*}}\ell = \frac{\partial \ell (y_i, f^{m^*}(x_i^{m^*}))}{\partial \theta^{m^*}}$\; 
        \textit{// Update parameters of the winner network}\;
        $\theta^{m^*} \gets \theta^{m^*} - \eta \nabla_{\theta^{m^*}}\ell$ \;
        \KwRet $\theta^{m^*}$\;
        }
        \vspace{0.5em}
  \Fn{\FLoser{$\theta^{m^c}$, $x_i^{m^c}$, $y_i$, $f$}}{
     \textit{// Compute soft targets of $f^{m^*}$ using Eq. \ref{eq:soft_targ}}\;
     $s_i^{m^*} = \sigma(f^{m^*}_i(x_i^{m^*}) / T)$\;
     \textit{// Compute soft targets of $f^{m^c}$ using Eq. \ref{eq:soft_targ}}\;
     $s_i^{m^c} = \sigma(f^{m^c}_i(x_i^{m^c}) / T)$\;
     \textit{// Compute the gradient w.r.t. GD loss using Eq. \ref{eq:loss_GD}}\;
     $\nabla_{\theta^{m^c}}\ell^{GD} = \frac{\partial \ell^{GD} (y_i, f^{m^c}, s_i^{m^*}, s_i^{m^c})}{ \partial \theta^{m^c}}$\;
  \textit{// Update parameters of the loser networks}\;
     $\theta^{m^c} \gets~ \theta^{m^c} - \eta \nabla_{\theta^{m^c}}\ell^{GD}$ \;
     \KwRet $\theta^{m^c}$\;
  }
\caption{DMCL}
\label{alg:gmcl}
\end{algorithm}

\subsection{Relationship to other MCL methods}
The general framework for MCL is described in lines 1-17 of Algorithm \ref{alg:gmcl}. The main idea is to enable each of the networks of the ensemble to specialize in different parts of the problem. This algorithm was first devised for RGB ensembles. 
Two recent instances of MCL are Stochastic MCL (SMCL) \cite{lee2016stochastic} and Confident MCL (CMCL) \cite{lee2017confident}. These methods differentiate from each other and from the general MCL framework in two fundamental ways: 1) the criterion loss used to decide whether a network is a winner or a loser (line 10, Algorithm~\ref{alg:gmcl}), and 2) how winner and loser models are updated (line 11 and 14, Algorithm~\ref{alg:gmcl}).
In SMCL, $\ell_{criterion}$ corresponds to the task loss, \eg standard cross-entropy for classification. The winner model is updated with respect to that same loss, while the loser models are not updated. This update scheme is also used in \cite{tian2019versatile}.
In CMCL, the $\ell_{criterion}$ corresponds to the task loss plus an additional loss that measures how well the other networks predict the uniform distribution, for the given sample. The winner model is updated as in the SMCL method and the loser models are updated with respect to the $KL$ divergence between its predictions and the uniform distribution.

Neither variations of MCL satisfy our problem statement. SMCL does not result in a single prediction. While CMCL does result in a single prediction by averaging the predictions, it does not account for the idiosyncrasies of multimodal data. The first aspect has to do with heterogeneous training dynamics resulting from having multimodal data as input. Figure \ref{fig:train-speeds} shows the cross-entropy loss of three networks independently trained for action recognition, using RGB (blue), optical flow (orange), and depth (green). Optical flow learns at a much faster speed than the other modalities. This results in an undesired effect when using CMCL: the optical flow network repeatedly achieves the lowest loss. This behavior is reinforced by the \textit{argmin} operator and the update scheme of CMCL, that does not allow useful gradients to pass to the loser networks. Eventually, the optical flow network ends up winning for all the training samples, which renders the other networks and modalities useless. The second challenge is the probable overfitting. The current training update scheme dictates that only the winner network gets useful gradients to build good representations for the given task, which reduces the data used to train each network. To address this and prevent overfitting, CMCL proposes to share the lower layers of the feature encoders. This is not feasible when the different networks are learning from different modalities as their representations/domains are significantly different. 

DMCL addresses these issues for multimodal data by using a cooperative learning setting where the ensemble networks teach each other via Knowledge Distillation. At the same time, DMCL leverages the ensemble learning strategy of the traditional MCL framework, where models specialize depending on their performance with respect to a given input. 

\section{Experiments}

In this section, we present the action recognition benchmark datasets we use to evaluate our approach. We then present the architecture and setup of our experiments. We analyze the performance of our DMCL in comparison to other MCL training strategies. We give insight into why other MCL training strategies fall short for multimodal data. We then demonstrate our privileged information state-of-the-art results and conclude with a discussion of our experimental results.

\subsection{Datasets} 
We test DMCL on three video action recognition datasets that offer RGB and depth data. We augment the three datasets with optical flow frames obtained using the implementation available at \cite{pyflow}, based on Liu \etal \cite{liu2009beyond}.

\textbf{Northwestern-UCLA (NW-UCLA).} This dataset \cite{wang2014cross} features ten people performing ten actions, captured simultaneously at three different viewpoints. We follow the cross-view protocol suggested by the authors in \cite{wang2014cross}, using two views for training and the remaining for testing. 

\textbf{UWA3DII.} This dataset \cite{rahmani2016histogram} features ten subjects performing thirty actions for four different trials, each trial corresponding to a different viewpoint. As suggested in \cite{rahmani2016histogram}, we follow the cross-view protocol using two views for training and two for testing.

\textbf{NTU120.} The very recent NTU RGB+D 120 dataset \cite{shahroudy2016ntu} is one of the largest multimodal dataset for video action recognition. It consists of a total of 114,480 trimmed video clips of 106 subjects performing 120 classes, including single person and two-person actions, across 155 different viewpoints and 96 background scenes. We follow the cross-subject evaluation protocol proposed in the original paper, using fifty three subjects for training and the remaining for testing. We also create three versions of NTU120, which we refer to as NTU120$^{mini}$, that contains 50\% sampled training data from the 120 classes. We note that NTU120 and NTU120$^{mini}$ share the same test data. When results are reported on NTU120$^{mini}$ they are averaged over the three runs.  We also evaluate our method on the smaller less recent version of this dataset, NTU60 \cite{shahroudy2016ntu}, that has 60 classes, in order to compare against state-of-the-art reported results.

\begin{figure}
    \begin{center}
    \includegraphics[width=0.9\linewidth,trim={0.4in 0.1in 0.6in 0.3in},clip]{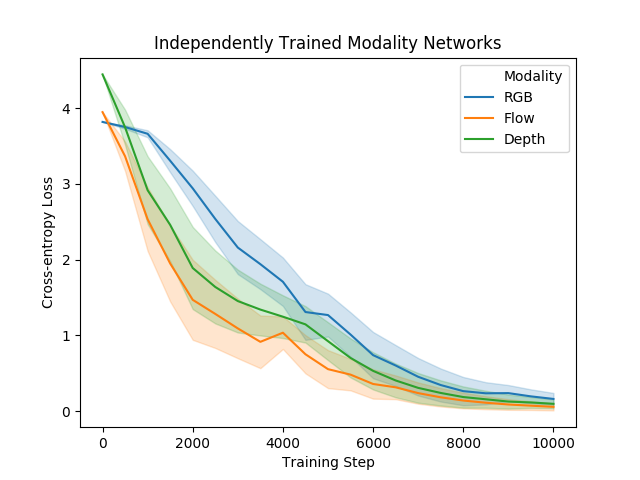}
    \end{center}
    \vspace{-4mm}
  \caption{\footnotesize The cross-entropy loss of three networks independently trained for action recognition on the UWA3DII dataset, using RGB (blue), depth (green), and optical flow (orange). These plots are averaged over three runs. We observe that for the first 10K steps, the training loss of the optical flow network is consistently lower, resulting in a winner-takes-all behavior in traditional MCL algorithms. However, in DMCL, the winner network also teaches the loser networks, strengthening the other modality networks and avoiding this behavior.}
  \label{fig:train-speeds}
\end{figure}

\subsection{Architecture and Setup}
Each modality network is implemented as the \FuncSty{R(2+1)D-18} architecture proposed in \cite{tran2018closer}.
This architecture is based on a Resnet-18 network \cite{he2016deep}, modified such that a 1D temporal convolution is added after every 2D convolution, thus giving the network the ability to learn spatiotemporal features. The factorization of a 3D convolution into a combination of 2D + 1D convolution has shown to be more effective for video classification tasks. The ensemble of modality networks is simultaneously trained following Algorithm~\ref{alg:gmcl}.

\begin{table*}
\small
\begin{center}
\begin{tabular}{l|ccc|ccc|ccc|ccc}
\toprule
& \multicolumn{3}{c|}{\textbf{Independent}} & \multicolumn{3}{c|}{\textbf{SMCL}} & \multicolumn{3}{c|}{\textbf{CMCL}} & \multicolumn{3}{c}{\textbf{Our DMCL}} \\
& RGB & $\sum$ & $\Phi$ & RGB & $\sum$ & $\Phi$ & RGB & $\sum$ & $\Phi$ & RGB & $\sum$ & $\Phi$ \\
\midrule
NWUCLA & 87.53 & 93.79 & 97.86 & 24.83 & 49.00 & 86.79 & 11.13 & 84.73 & 89.65 & \boldrgb{93.64} & 93.28 & 97.64 \\
UWA3DII & 73.74 & 89.75 & 95.52 & 25.19 & 60.70 & 88.51 & 22.28 & 31.90 & 83.89 & \boldrgb{78.39} & 89.50 & 94.96 \\
NTU120$^{mini}$ & 79.66 & 86.57&92.11&26.67&62.22&86.19&29.61&5.28&86.29&\boldrgb{81.25}&86.23&91.71 \\
NTU120 & \boldrgb{84.86} & 89.74 & 94.36 & 22.31 & 5.54 & 79.81 & 22.37 & 5.06 & 85.20 & 84.31 & 88.46 & 93.21\\
\bottomrule
\end{tabular}
\end{center}
\vspace{-2mm}
\caption{\footnotesize \textbf{Comparing MCL methods.} We compare the performance of SMCL and CMCL with our proposed DMCL on the NWUCLA, UWA3DII, and NTU120 datasets. We also compare against independently trained modality networks. 
For each method we present the accuracy of the RGB modality network, the sum of all modality network predictions ($\sum$), and the oracle accuracy ($\Phi$). For each row, corresponding to one dataset, we highlight in bold the best result using RGB only at test time. Using our DMCL methods results in better RGB networks for three out of four datasets.}
\label{tab:compare-mcl}
\end{table*}

The input of each modality network is a clip of eight frames of the corresponding modality. For each training step, a video is split into eight equal parts and we randomly sample a frame from each of them. 
Each training input frame is a crop of dimension [224,224,3], cropped around a randomly shifted center, for each video. We also use other data augmentation techniques such as random horizontal flipping and random color distortions. 
The networks are trained from scratch for all the experiments, using SGD optimizer with Momentum $0.9$, and an initial learning rate of $10^-3$.
At test time, we sample ten clips per video, each clip consisting of eight frames randomly sampled, centered, and with no data augmentation techniques. The final prediction for each video is the average of the ten clip predictions. We have experimented with different values of temperature $T$ and hyperparameter $\lambda$, and found that $T$=\{2,5\} and $\lambda$=\{1, 0.5\} works best, with little accuracy variations. Further details related to hyperparameters are given in the supplementary material.

\subsection{Results}
In this section, we demonstrate how DMCL leverages multiple modalities to learn an RGB network that outperforms an independently trained RGB classifier - our baseline, and other MCL training strategies. All MCL strategies are trained using the same training process as our method, including data augmentation techniques, optimizer, and number of steps, and are considered as ablation experiments of our method. We then demonstrate state-of-the-art privileged information results.

\smallskip

\textbf{Comparison vs. MCL variants.} Table \ref{tab:compare-mcl} shows the action classification performance on the three video action recognition benchmark datasets for MCL variants and independently trained modality networks. We present the classification accuracy using the RGB modality, the sum of predictions of RGB, Flow, and Depth modalities ($\Sigma$), and the oracle accuracy ($\Phi$). An oracle $\Phi$ is assumed to have the ability to select the modality that gives the best prediction among the ensemble. Our DMCL approach performs better than modalities trained independently, \ie without MCL, and better than SMCL and CMCL variants.  While Table \ref{tab:compare-mcl} focuses on improvement with regard to the RGB modality, we provide similar results for Depth and Optical Flow in the supplementary material. We note that the effect of knowledge distillation is more visible in the three smaller datasets.

\begin{table}
\small
\begin{center}
\begin{tabular}{r|ccccc}
\toprule
 \multicolumn{6}{c}{\textbf{KNN accuracy with random features}}\\
 \textbf{Modality} & $k$=1 & $k$=5 & $k$=10 & $k$=50 & $k$=120\\
\midrule
RGB &10.53 &10.74 &11.11 &11.32 &12.26  \\
Depth &9.72 &10.68 &10.77 &15.37 &13.31 \\
\textbf{Optical Flow} & 23.23& 23.96&25.31 &26.35 &24.53 \\
\bottomrule
\end{tabular}
\end{center}
\vspace{-2mm}
\caption{\footnotesize Accuracy of a KNN classifier with varying $k$ on the NWUCLA dataset. Classified features are computed using randomly initialized networks for each modality. Although all features are randomly generated, optical flow random features tend to achieve a significantly higher accuracy. This helps to explain why optical flow networks learn faster than other modalities.}
\label{tab:knn}
\end{table}

Table \ref{tab:compare-mcl} also shows that combining the predictions of three modalities ($\Sigma$) generally improves accuracy. The fact that the oracle accuracy ($\Phi$) is significantly higher than $\Sigma$ indicates that, for some cases, at least one modality predicted the correct class, however, the sum of predictions ($\Sigma$) resulted in an incorrect prediction. However, the gap between $\Sigma$ and $\Phi$ is lower for DMCL compared to the other approaches. This indicates that DMCL combines modality predictions in a more optimal fashion to improve overall accuracy. The low accuracies of SMCL and CMCL are due to artifacts created by the use of multimodal data, which we investigate in the next section. We have checked the implementation of these methods on RGB-only ensembles, which lead to similar results to those reported in the original papers.

\begin{table*}[h]
\small
\begin{center}
\begin{tabular}{r|ccccc|ccccc}
\toprule
Dataset & \multicolumn{5}{c}{\textbf{NWUCLA}} & \multicolumn{5}{c}{\textbf{UWA3DII}}\\
Test Modality & RGB & Depth & Flow & $\Sigma$ & $\Phi$ & RGB & Depth & Flow & $\Sigma$ & $\Phi$ \\
\midrule
Independent & 87.53& 80.30& 89.58& 93.79 & 97.86 & 73.74 & 77.09& \boldflow{89.66} & 89.75 & 95.52\\
Random Teacher & 89.57 & 57.81& 89.43& 86.93 & 95.71 & 71.07& 79.07& 85.03 & 84.47 & 92.60\\
\textbf{Our DMCL} & \boldrgb{93.64} & \bolddepth{83.29}& \boldflow{91.07} & 93.28 & 97.64 & \boldrgb{78.39}& \bolddepth{81.87}& 88.26  & 88.51 & 94.59\\
\bottomrule
\end{tabular}
\end{center}
\vspace{-2mm}
\caption{\footnotesize \textbf{Selecting the right teacher network is important.} We present the action recognition classification accuracy on the NWUCLA and UWA3DII datasets for three scenarios, where: modality networks are trained independently; a random teacher is assigned for every sample to guide the other modality networks; and DMCL, where the best-performing teacher (lowest loss) is selected to guide other modality networks. For each column, corresponding to a test modality, we highlight in bold the best result across the three scenarios.}
\label{tab:rndT}
\end{table*}

\smallskip

\textbf{Learning speed for different modalities.} One of the goals of this paper is to investigate and bring new insights on multimodal learning.
In a MCL setting, having a specific modality learn at a faster pace compared to others often leads to an imbalance of the number of data points each modality network is presented with at training time.
Networks specializing in different modalities typically do not share a backbone of parameters due to the very different nature of the inputs - in contrast to the SMCL and CMCL variants where there is a shared backbone. As a consequence, if a modality network dominates the training process, \ie being the one to consistently achieve the lowest loss for training batches, it will be presented with significantly more training data compared to the other modality networks.
We observed that optical flow often dominates the ensemble training process particularly when training using CMCL. This is depicted in Figure~\ref{fig:train-speeds} where the training loss curves of the independently trained networks for Optical Flow, Depth, and RGB are shown over the training steps.
Namely, looking at the first steps of the curve we see that Optical Flow curve is consistently lower than Depth, which in turn has lower values than RGB. This is consistent with what we find during training of CMCL, where the RGB network is often ignored, the Depth network learns from a few samples and overfits early, and the Optical Flow network sees the vast majority of the samples.

We further investigate why optical flow dominates the learning process in our action recognition setting. We compute random features extracted from a randomly initialized untrained network for each of the modalities using the same architecture described previously. We then run a $k$NN classifier using the random features. Table \ref{tab:knn} shows results of this experiment on the NWUCLA dataset for $k=1,5,10,50,120$. The accuracy of the random features of the optical flow modality is almost twice that  achieved using Depth and RGB. The fact that the $k$NN classifier achieves such good performance compared to the other modalities suggests that Optical Flow data naturally clusters better per class. From the perspective of a deep neural network learning process, this could be interpreted as a better initialization, thus speeding the initial stage of learning.

\smallskip

\textbf{Leveraging Teacher Strength.} In this section, we ablate the mechanism by which the teacher role is determined. The teacher role is assigned to the network that achieves the lowest loss for each sample of the batch, therefore being in the best position to guide/strengthen the other networks. To verify this claim, we train our model with a random assignment of a  teacher for each sample of the batch. This can be though of as a randomized distillation process. We then compare the overall action recognition classification accuracy of both approaches in Table \ref{tab:rndT}. Choosing the right network as teacher consistently achieves better performance compared to a randomly assigned teacher, for every modality. This is in-line with work that combines distillation and graphs, where the distillation process has a specific direction specified by the direction of the edges \cite{Luo_2018_ECCV}.
It is interesting to note that random teacher assignment may result in better performance than individual modality networks, \eg for NWUCLA the RGB individual network accuracy is 87.53\% \vs 89.57\% for a random teacher assignment. These may be related to the known regularization effect of knowledge distillation, that has been empirically shown to lead to better performance \cite{hinton2015distilling,furlanello2018born}.

\smallskip

\textbf{State-of-the-art Comparisons.} We now compare DMCL to state-of-the-art privileged information methods, and modality baselines, for the task of human action recognition from videos. Table \ref{tab:sota-small} shows results for the UWA3DII and NWUCLA datasets. 
The top part of the table presents modality baselines for methods that use the same number of modalities in training and testing, including our individually trained modality networks. The bottom part of the table refers to methods that have missing modalities at test time.
Our DMCL using RGB only for testing achieves higher accuracy compared to all baselines that use RGB at training and testing, and compared to all state-of-the-art privileged information methods that use RGB at test time, including those that use additional hallucination networks at test time, achieving an absolute improvement of 4.7\% for UWA3DII and 6.1\% for NWUCLA. Similarly, our DMCL outperforms all baselines when the only available modality is Depth by 4.8\% absolute improvement and the state-of-the-art method by 1.3\% on UWA3DII.

Table \ref{tab:sota-ntu120} presents results on three versions of the NTU dataset: NTU60, NTU120$^{mini}$, and the full NTU120. We see that the distillation effect is much more visible in the case of less data. For example, for NTU$^{mini}$, we achieve an absolute improvement of 1.6\% over the baseline for the RGB modality, and of 6\% for NTU60. Our best modality network for NTU60 achieves 85.65\% compared to the 89.5\% of \cite{Luo_2018_ECCV} that uses twice the number of modalities we use for training and an additional graph network module.

\begin{table*}[h!]
\small
\begin{center}
\begin{tabular}{clcccc}
\toprule
& \textbf{Method} & \textbf{Training Modalities} & \textbf{Testing Modalities} & \textbf{UWA3DII} & \textbf{NWUCLA} \\
\midrule
\multirow{8}{*}{\rotatebox[origin=c]{90}{\textbf{\textit{\scriptsize{Modality Baselines}}}\hspace*{0.2em}}} & R-NKTM \cite{rahmani2018learning} & Syn* & RGB  & 66.3 & 78.1\\
& Action Tubes \cite{gkioxari2015finding} & RGB & RGB & 33.7 & 61.5\\
& Long-term RCNN \cite{donahue2015long} & RGB & RGB & \rgb{74.5} & 64.7\\ 
& Baseline (RGB)  & RGB & RGB  & 73.74 & \rgb{87.52} \\
&  MVDI+CNN \cite{xiao2019action} & Depth  & Depth & 68.3 & \bolddepth{84.2}\\
&   Baseline (D) & Depth & Depth  & \depth{77.09} & 80.30\\
&   Baseline (F) & Flow & Flow  & \boldflow{89.66} & \flow{89.58}\\
&   Baseline (RGB, D, F) & RGB, Depth, Flow & RGB, Depth, Flow  & 89.75 & 93.9\\
\midrule
\multirow{6}{*}{\rotatebox[origin=c]{90}{\textbf{\textit{\scriptsize{Privileged Information}}}\hspace*{0.2em}}} & Hoffman \textit{et al.} \cite{hoffman2016learning} & RGB, Depth  & RGB$^+$ &  66.67 & 83.30 \\
 & Garcia \textit{et al.} \cite{Garcia_2018_ECCV}  & RGB, Depth  & RGB$^+$ &  73.23 & 86.72\\
 & ADMD \cite{garcia2018learning}  & RGB, Depth  & RGB$^+$ & - & 91.64\\
 & \textbf{DMCL}  & RGB, Depth, Flow & RGB  & \boldrgb{78.39} & \boldrgb{93.64} \\
 & \textbf{DMCL}  & RGB, Depth, Flow & Depth  & \bolddepth{81.87} & \depth{83.29} \\
 & \textbf{DMCL}  & RGB, Depth, Flow & Flow  & \flow{88.26} & \boldflow{91.07}\\
\bottomrule
\end{tabular}
\end{center}
\vspace{-2mm}
\caption{\footnotesize \textbf{Accuracy for UWA3DII and NWUCLA dataset.} The first part of the table refers to methods that use unsupervised feature learning (*) or that use the same number of modalities for training and testing. The second part of the table refers to methods that use more modalities for training than for testing. Methods that use RGB$^+$ at test time use an additional network that mimics the missing modality. For each column, corresponding to one dataset, we highlight in colored bold the best result and in normal colored font the second best between our method and the baselines. Each color corresponds to a different test modality. To conduct a fair comparison with baseline methods, this table presents results for the most common view setting for UWA3DII and NWUCLA. Other view settings follow the same trend and results are presented in the supplementary material.}
\label{tab:sota-small}
\end{table*}

\begin{table*}[h!]
\small
\begin{center}
\begin{tabular}{clccccc}
\toprule
& \textbf{Method} & \textbf{Training Modalities} & \textbf{Testing Modalities} & \textbf{NTU60} & \textbf{NTU120$^{mini}$} & \textbf{NTU120} \\
\midrule
\multirow{9}{*}{\rotatebox[origin=c]{90}{\textbf{\textit{\scriptsize{Modality Baselines}}}\hspace*{0.2em}}} & ST-LSTM \cite{liu2016spatio}\cite{liu2019ntu} & Skeleton & Skeleton & 69.2 & $\sim$ 50.0 & 55.7  \\
& VGG \cite{liu2019ntu} & RGB & RGB & -& $\sim$ 40.0 & 58.5 \\
& Baseline (RGB) & RGB & RGB & \rgb{77.59} & \rgb{79.66} & \boldrgb{84.86} \\
& VGG \cite{liu2019ntu} & Depth & Depth & -& $\sim$ 20.0 & 48.7 \\
& Baseline (D) & Depth & Depth & \depth{78.97} & \depth{78.67} & \bolddepth{83.32} \\
& Baseline (F) & Flow & Flow & \flow{81.43} & \flow{84.21} & \boldflow{86.72} \\
& VGG \cite{liu2019ntu} & RGB,Depth & RGB, Depth  & -& - & 61.9 \\
& VGG \cite{liu2019ntu} & RGB, Depth, 3D Skeleton  & RGB, Depth, 3D Skeleton & - & - & 64.0 \\
& Baseline (RGB, D, F) & RGB, Depth, Flow & RGB, Depth, Flow & 87.25 &86.57& 89.74 \\
\midrule
\multirow{6}{*}{\rotatebox[origin=c]{90}{\textbf{\textit{\scriptsize{Privileged Information}}}\hspace*{0.2em}}} & Garcia \textit{et al.} \cite{garcia2018learning}  & RGB, depth  & RGB &  73.11 & - & -\\
& ADMD \cite{Garcia_2018_ECCV}  & RGB, Depth  & RGB &  73.4 & - & -  \\
& Luo \textit{et al.} \cite{Luo_2018_ECCV} & RGB, OF, Depth, 3D Skeleton$^{1,2,3}$  & RGB & \boldrgb{89.5} & - & -\\
& \textbf{DMCL} & RGB, Depth, Flow & RGB  & 83.61 & \boldrgb{81.25} & \rgb{84.31} \\
& \textbf{DMCL} & RGB, Depth, Flow & Depth & \bolddepth{80.56} & \bolddepth{78.98} &  \depth{82.22} \\
& \textbf{DMCL} & RGB, Depth, Flow & Flow  & \boldflow{85.65} & \boldflow{84.45} & \flow{86.44} \\
\bottomrule
\end{tabular}
\end{center}
\vspace{-2mm}
\caption{\footnotesize \textbf{NTU Datasets} The test sets for NTU120$^{mini}$ and NTU120 are the same. For each column, corresponding to one dataset, we highlight in bold the best result and in normal colored font the second best between our method and the baselines. Each color corresponds to a different test modality. The approximated values are inferred from a plot in \cite{liu2019ntu}. We note that the effect of the distillation method is more visible on the smaller scale versions NTU60 and NTU120$^{mini}$ of the dataset.}
\label{tab:sota-ntu120}
\end{table*}

\vspace{-2mm}
\section{Conclusions}
\vspace{-1mm}
MCL is a powerful way for training ensembles of networks, originally proposed for RGB data. We demonstrate undesirable behaviors of this framework when naively applied to multimodal data. 
We propose DMCL that extends MCL frameworks to leverage the complementary information offered by the multimodal data to the benefit of the ensemble. The cooperative learning is enabled via knowledge distillation that allows the ensemble networks to exchange information and learn from each other. We demonstrate that modality networks trained using our DMCL achieve competitive to or state-of-the-art results compared to the privileged information literature, and significantly higher accuracy compared to independently trained modality networks for human action recognition in videos.


{\small
\bibliographystyle{ieee_fullname}
\bibliography{egbib}
}

\end{document}